\documentclass[10pt,twocolumn]{article}

\usepackage{iccv}
\usepackage{times}
\usepackage{graphicx}
\usepackage{amsmath}
\usepackage{amssymb}
\usepackage{booktabs}
\usepackage{enumitem}
\usepackage{microtype}
\usepackage[export]{adjustbox}
\usepackage{nicefrac}
\usepackage{mathtools}
\usepackage{caption}
\usepackage{subcaption}

\usepackage[pagebackref=true,breaklinks=true,letterpaper=true,colorlinks,bookmarks=false]{hyperref}

\usepackage[capitalize]{cleveref}
\crefname{section}{Sec.}{Secs.}
\Crefname{section}{Section}{Sections}
\Crefname{table}{Table}{Tables}
\crefname{table}{Tab.}{Tabs.}

\iccvfinalcopy 

\makeatletter
\DeclareRobustCommand\onedot{\futurelet\@let@token\@onedot}
\def\@onedot{\ifx\@let@token.\else.\null\fi\xspace}
 
\def\ie{\emph{i.e}\onedot}

\makeatother

\newcommand{\citesupmat}[1]{%
  \if@filesw
    \protected@write\@auxout{}{\string\citation{#1}}%
  \fi
  [\cite{#1}]%
}

  \def\cL{{\mathcal L}}

\newcommand{\img}{\mathbf{x}}
\newcommand{\rec}{\hat{\img}}
\newcommand{\layer}{\mathbf{o}}
\newcommand{\layerc}{\layer^c}
\newcommand{\layera}{\layer^\alpha}
\newcommand{\scode}{z}
\newcommand{\pcode}{z^p}
\newcommand{\proba}{p}
\newcommand{\sprite}{\mathbf{s}}

\newcommand{\feature}{f}
\newcommand{\encoder}{e_\theta}
\newcommand{\generator}{g_\theta}
\newcommand{\bkg}{b_\theta}

\newcommand{\lrec}{\cL_{\textrm{rec}}}

\newcommand{\lsup}{\cL_{\textrm{sup}}}

\newcommand{\lctc}{\cL_{\textrm{ctc}}}
\newcommand{\lambdactc}{\lambda_\textrm{ctc}}
\newcommand{\gt}{y}
\newcommand{\pred}{\hat{\gt}}

\DeclareMathOperator*{\softmax}{softmax}
\DeclareMathOperator*{\dotprod}{\cdot}

\begin{document}

\title{The Learnable Typewriter: A Generative Approach to Text Analysis}

\author{{Ioannis Siglidis \;\; Nicolas Gonthier \;\; Julien Gaubil \;\; Tom Monnier \;\; Mathieu Aubry }\\
{LIGM, Ecole des Ponts, Univ Gustave Eiffel, CNRS, Marne-la-Vallée, France}\\
\tt\small{\url{https://imagine.enpc.fr/~siglidii/learnable-typewriter}}
}
\maketitle

\begin{abstract}
    We present a generative document-specific approach to character analysis and recognition in text lines. Our main idea is to build on unsupervised multi-object segmentation methods and in particular those that reconstruct images based on a limited amount of visual elements, called sprites. Taking as input a set of text lines with similar font or handwriting, our approach can learn a large number of different characters and leverage line-level annotations when available. Our {contribution is} twofold. First, we provide the first adaptation and evaluation of a deep unsupervised multi-object segmentation approach for text line analysis. Since these methods have mainly been evaluated on synthetic data in a completely unsupervised setting, demonstrating that they can be adapted and quantitatively evaluated on real images of text and that they can be trained using weak supervision are significant progresses. Second, we show the potential of our method for new applications, more specifically in the field of paleography, which studies the history and variations of handwriting, and for cipher analysis. We demonstrate our approach on three very different datasets: a printed volume of the Google1000 dataset~\cite{vincent_google_2007,gupta_learning_2018}, the Copiale cipher~\cite{baro_generic_2019,knight_copiale_2011} and historical handwritten charters from the 12th and early 13th century~\cite{dh2022,pal}.  
\end{abstract}
\section{Introduction}
\label{sec:intro}

\begin{figure}
  \centering
  \begin{subfigure}{1.0\linewidth}
    \includegraphics[width=1.0\textwidth]{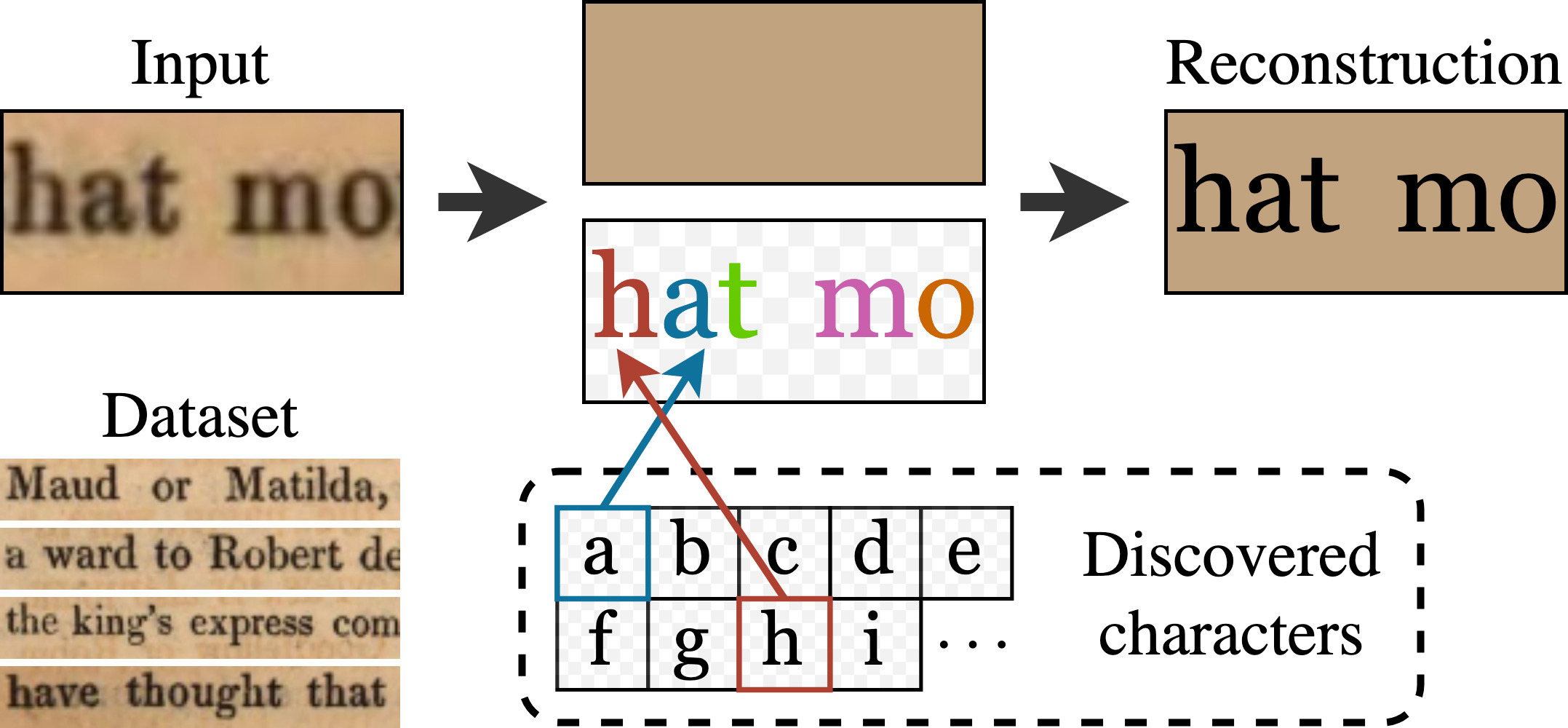}
    \caption{The Learnable Typewriter idea}
    \vspace{.5em}
    \label{fig:teaser_overview}
  \end{subfigure}

  \begin{subfigure}{0.45\linewidth}
    \includegraphics[width=\textwidth]{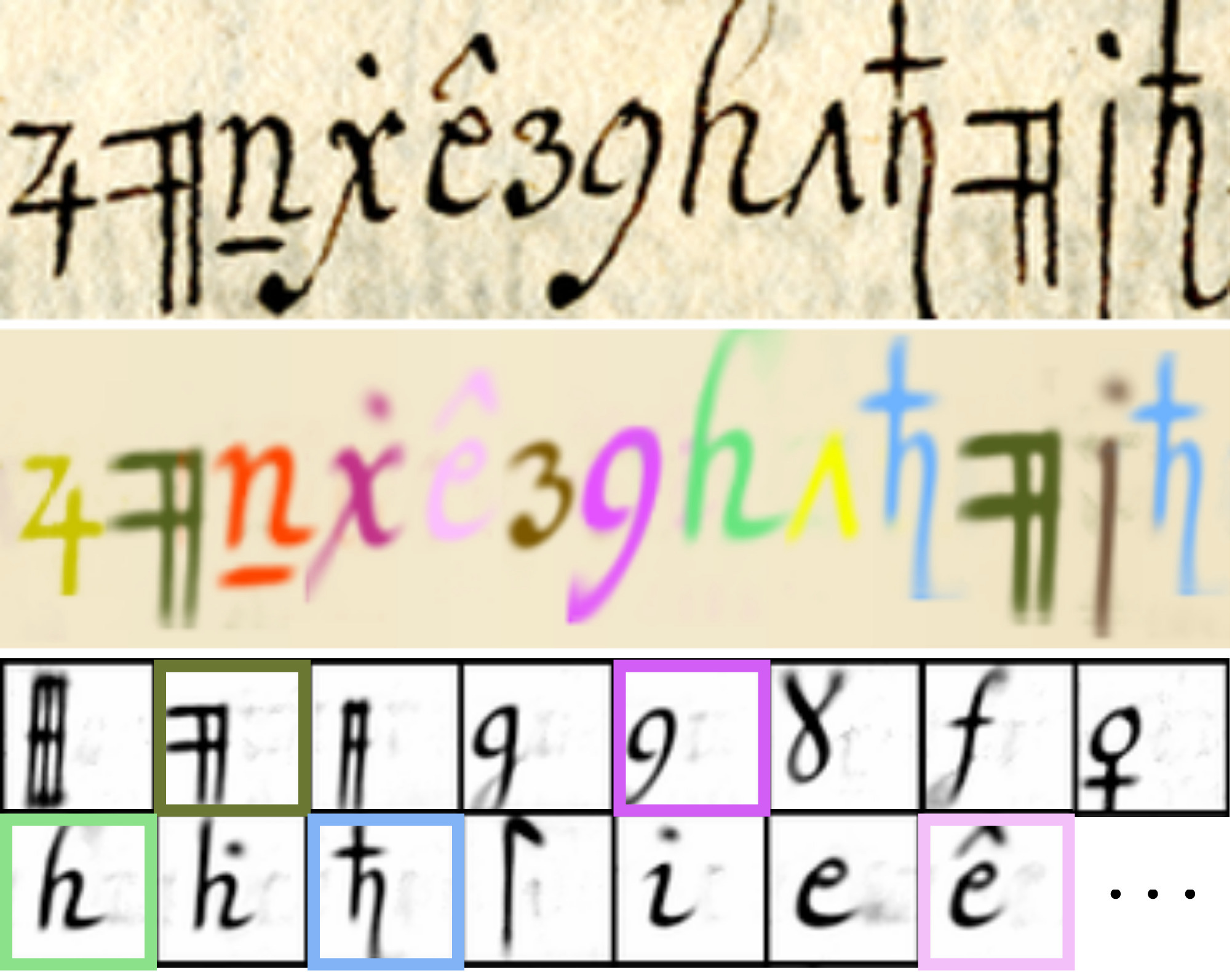}
    \caption{Cipher analysis~\cite{knight_copiale_2011}}
    \label{fig:teaser_copiale}
  \end{subfigure}
  \hfill
  \begin{subfigure}{0.50\linewidth}
  \includegraphics[width=\textwidth]{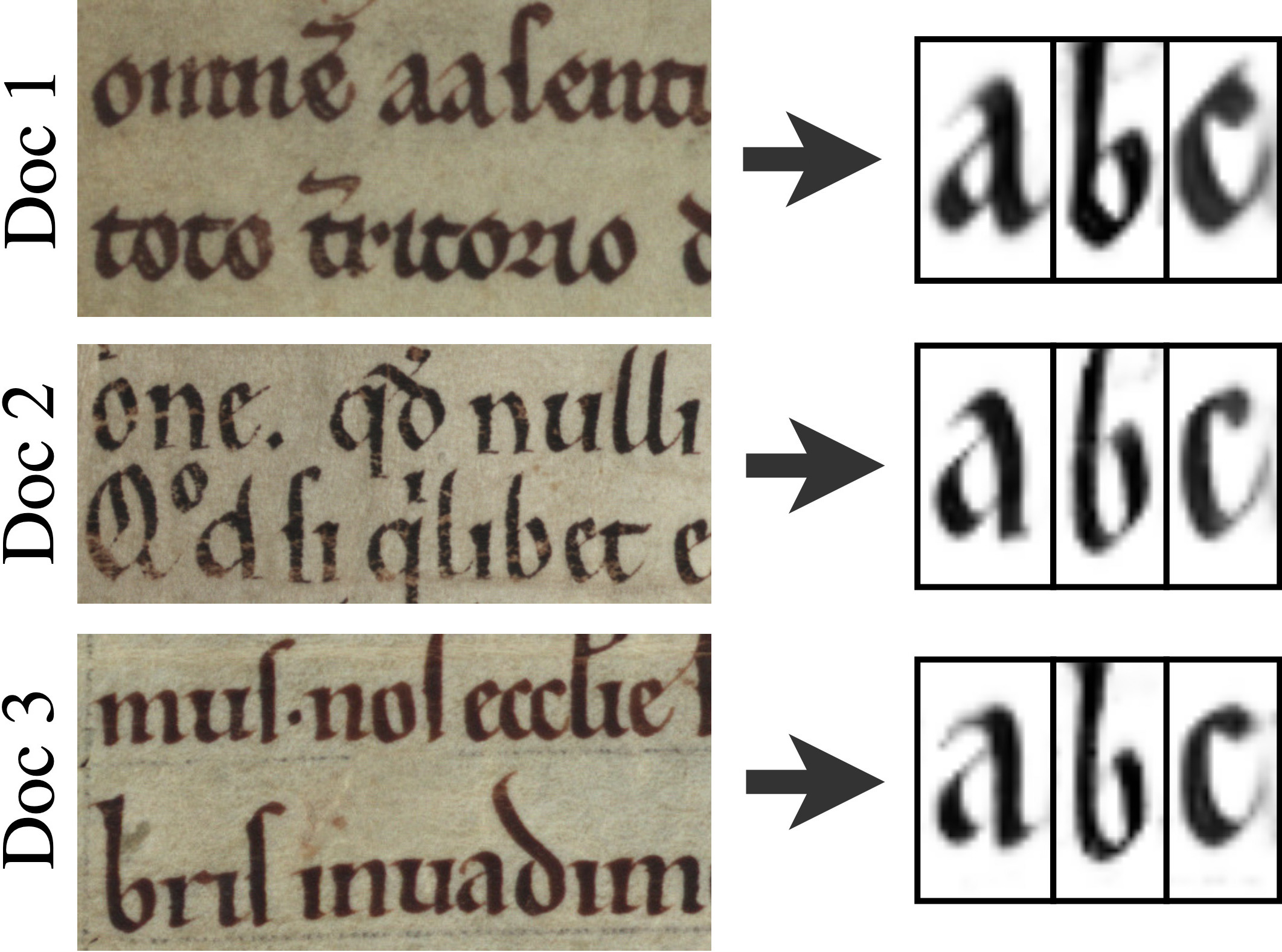}
  \caption{Paleographic analysis~\cite{pal} }
  \label{fig:teaser_paleo}
  \end{subfigure}  
  \caption{\textbf{The Learnable Typewriter. (a)} Given a text line dataset, we learn to reconstruct images to discover the underlying characters. Such a generative approach can be used to analyze complex ciphers \textbf{(b)} and can be used as an automatic tool to help the study of handwriting variations in historical documents \textbf{(c)}.}
  \label{fig:teaser}
\end{figure}
 A popular approach to document analysis in the 1990s was to learn document-specific character prototypes, which enabled Optical Character Recognition (OCR)~\cite{kopec1996document,kopec1997supervised,xu1999prototype,baird1999model} but also other applications, such as font classification~\cite{hochberg1997automatic} or document image compression and rendering~\cite{nolan2010method}. This idea culminated in 2013, with the Ocular system~\cite{berg-kirkpatrick_unsupervised_2013} which proposed a generative model for printed text lines inspired by the printing process and held the promise of achieving a complete explanation of their appearance. These document-specific generative approaches were however overshadowed by discriminative approaches, whose sole purpose is to perform predictions, and which lead to higher performance at the cost of interpretability, e.g.~\cite{graves2008offline,li2021trocr}. In this paper, we explore how modern deep approaches enable revisiting and extending model-based approaches to text line analysis. In particular, we demonstrate an approach that can deal with challenging examples of handwritten documents, opening a new perspective for the study of historical handwriting, paleography.

 While discriminative approaches are largely dominant in today's deep learning-based computer vision, a recent set of works revisited generative approaches for unsupervised multi-object object segmentation~\cite{burgess2019monet,emami2021efficient,greff2017neural,greff2019multi,yang2020learning,crawford2019spatially,deng2020generative,eslami_attend_2016,jiang2020generative,smirnov_marionette_2021,monnier_unsupervised_2021}. Most of them provide results on synthetic data or simple real images~\cite{monnier_unsupervised_2021}, and sometimes show qualitative results on simple printed text images~\cite{smirnov_marionette_2021,reddy2022search}. Surprisingly, images of handwritten characters, which
were notoriously used in the development of convolutional neural networks~\cite{lecun_backpropagation_1989,lecun_gradientbased_1998} and generative adversarial networks~\cite{goodfellow2014generative} were largely overlooked by these approaches. 

We build on recent sprite-based unsupervised image decomposition approaches \cite{smirnov_marionette_2021,monnier_unsupervised_2021} that provide an interpretable decomposition of images into a dictionary of visual elements, referred to as sprites. These methods jointly optimize both the sprites and the neural networks that predict their position and color. Intuitively, we would like to adapt these methods so that, from text lines that are extracted from any given document, they could learn sprites that correspond to each character. By adapting MarioNette~\cite{smirnov_marionette_2021} to perform text line analysis, we provide quantitative evaluation on real data and an analysis of the limitations of a state-of-the-art unsupervised multi-object segmentation approach. 
We argue that text line recognition  should be used as a benchmark for this task in future work. 

{Because unsupervised performances are not completely satisfactory, we combine this approach with a weak supervision from line-level transcriptions. Transcriptions are widely available and easy to produce with dedicated software, e.g.~\cite{kahle2017transkribus}, and we show this dramatically improves the results, while preserving their interpretability. We believe that similar weak (i.e., image-level) annotations could also be considered for other images decomposition problems.}
\noindent {\bf Contributions.} To summarize, we present:
\begin{itemize}[itemsep=0em, topsep=0em]%
    \item a deep generative approach to text line analysis, inspired by deep unsupervised multi-object segmentation approaches and adapted to work in both a {weakly supervised} and unsupervised setting,
    \item a demonstration of the potential of our approach in challenging applications, particularly {ciphered documents} 
    and paleographic analysis,
    \item {experiments} on three very different datasets:  a printed volume of the Google1000 dataset~\cite{vincent_google_2007,gupta_learning_2018}, the Copiale cipher~\cite{baro_generic_2019,knight_copiale_2011} and historical handwritten charters from the 12th and early 13th century~\cite{dh2022,pal}.
\end{itemize}
Our complete implementation can be found at \href{https://github.com/ysig/learnable-typewriter.git}{github.com/ysig/learnable-typewriter}.

\section{Related Work}


\begin{figure*}
  \centering
  \begin{subfigure}{0.46\linewidth}
  \centering
    \includegraphics[width=1.0\textwidth]{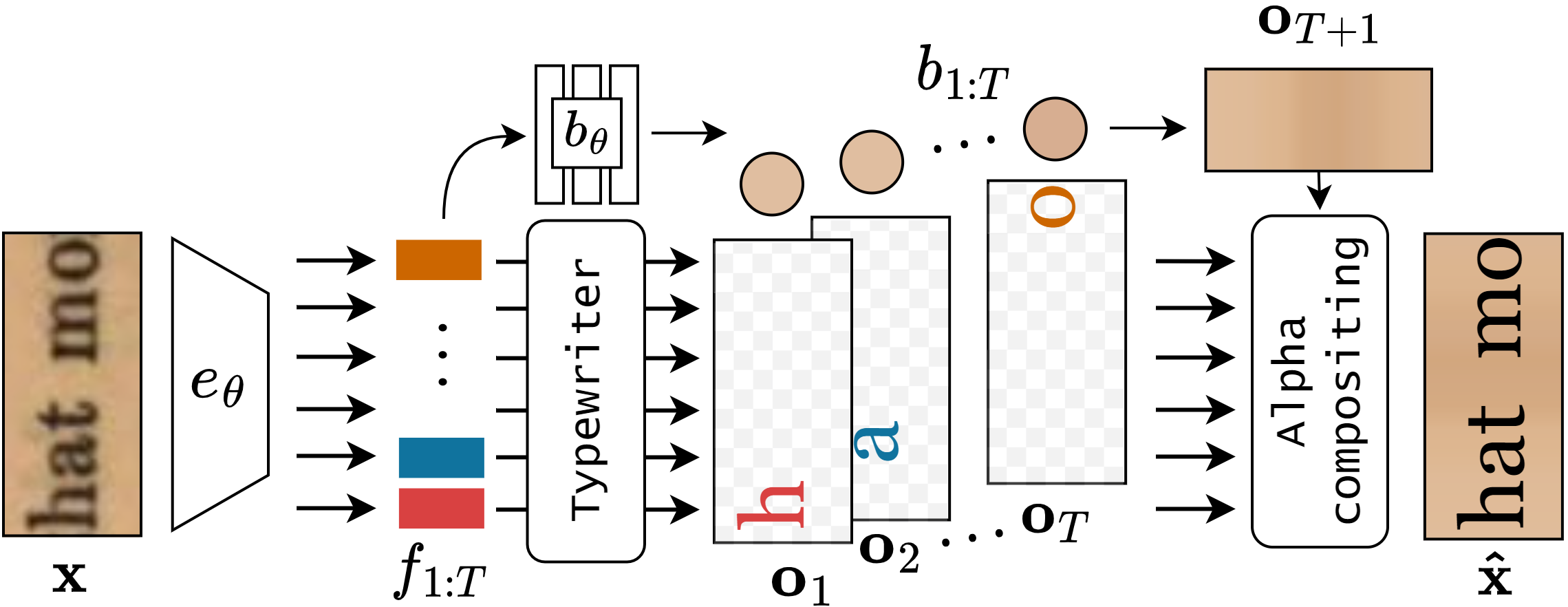}
    \caption{Overview of our full pipeline}
    \label{fig:overview}
  \end{subfigure}
  \hfill
  \begin{subfigure}{0.51\linewidth}
    \includegraphics[width=1\textwidth]{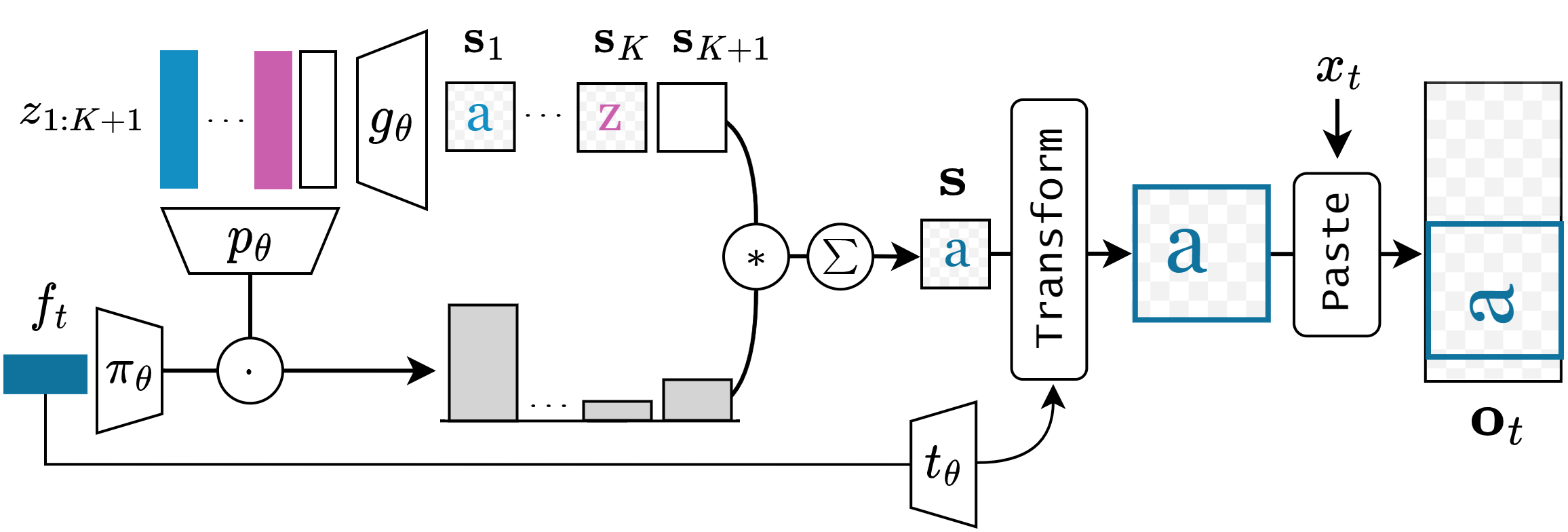}
    \caption{Details of our Typewriter module}
    \label{fig:writer}
  \end{subfigure}
  \caption{\textbf{Overview. (a)} An image is encoded into a sequence of features, each decoded by the Typewriter module into image layers. They are then fused by alpha compositing with a predicted uniform background. \textbf{(b)} The Typewriter module takes a feature as input, computes sprites and associated probabilities from learned latent codes, and composes them into a composite sprite that is transformed and positioned onto an image-sized canvas.} 
  \label{fig:pipeline}
  \vspace{-0.3em}
\end{figure*}
\paragraph{Text recognition.} 

Image Text Recognition, including Optical Character Recognition (OCR) and Handwritten Text Recognition (HTR), is a classic pattern recognition problem, and one of the earliest successful application of deep learning \cite{lecun_backpropagation_1989,lecun_gradientbased_1998}.
The mainstream approaches for text line recognition rely on discriminative supervised learning. Typically, a Convolutional Neural Network (CNN) encoder will map the input image to a sequence of features and a decoder will associate them to the ground truth, e.g. through a recurrent architecture trained with a Connectionist Temporal Classification (CTC) loss~\cite{ctc,graves2008offline,puig,bluche2017gated,de2020htr}, or a transformer trained with cross entropy~\cite{patwyr,li2021trocr}.

More related to our work, ScrabbleGAN~\cite{fogel_scrabblegan_2020} proposed a generative adversarial approach for semi-supervised text recognition, but their method is neither able to reconstruct an input text line nor to decompose it into individual characters. Also related are approaches which use already annotated sprites (referred to as exemplars or supports) to perform {OCR/HTR} \cite{zhang_adaptive_2020,HTRbyMatching} by matching them to text lines. Recent unsupervised approaches, either cluster input images in a feature space \cite{baro_generic_2019} or rely on an existing text corpus of the recognized language \cite{gupta_learning_2018}.

Closest to our work are classical prototype-based methods~\cite{kopec1996document,kopec1997supervised,xu1999prototype,baird1999model}  and in particular the Ocular system \cite{berg-kirkpatrick_unsupervised_2013} which follows a generative probabilistic approach to jointly model text and character fonts in binarized documents, and is optimized through Expectation Maximization (EM). Different from us, it also relies on a pre-trained n-gram language model, originally from the english language and later extended to multiple languages ~\cite{garrette_unsupervised_2015}. {Other} {approaches} {rely on language models to identify characters~\cite{kopec2001n,berg-kirkpatrick_unsupervised_2013,gupta_learning_2018}. However, language models do not exist for unknown ciphers, or historical manuscripts which are often strongly abbreviated. Instead, we propose to disambiguate sprites by relying on line level transcriptions.}


\paragraph{Unsupervised multi-object segmentation.} 

Unsupervised multi-object segmentation refers to a family of approaches that decompose and segment scenes into multiple objects in an unsupervised manner \cite{karazija_clevrtex_2021}. Some perform decomposition by computing pixel level segmentation masks over the whole input image \cite{burgess2019monet,emami2021efficient,greff2017neural,greff2019multi,yang2020learning}, while others focus on smaller regions of the input image and learn to compose objects in an iterative fashion, mostly using a recurrent architecture \cite{crawford2019spatially,deng2020generative,eslami_attend_2016,jiang2020generative}. 
All of these techniques can isolate objects by producing segmentation masks, but our goal is also to capture recurring visual elements.

We thus build on techniques that explicitly model the objects located inside the input image, by associating them to a set of image prototypes referred to as sprites~\cite{monnier_unsupervised_2021,smirnov_marionette_2021}. 
Sprites are color images with an additional transparency channel and are associated to transformation prediction networks that {are} used to compose them onto a target canvas. However, DTI-Sprites \cite{monnier_unsupervised_2021} can only predict a small amount of sprites for a collection of fixed-size images and fails to scale when the number of objects inside each image scales to those of real documents. At the same time,  MarioNette \cite{smirnov_marionette_2021} suffers from a high reconstruction error and fuzzy sprites that sub-optimally reconstruct a toy text dataset.

\section{The Learnable Typewriter}
\label{sec:Approach}
Given a collection of text lines written using consistent font or handwriting, our goal is to learn the shape of all the characters it contains and a deep network that predicts the exact way these characters were used to generate any input text line. Since complete supervision (\ie, the position and shape of every character used in a document) for such a task would be extremely costly to obtain, we propose to proceed in an analysis-by-synthesis fashion and to build on sprite-based unsupervised image decomposition approaches~\cite{smirnov_marionette_2021,monnier_unsupervised_2021} which jointly learn a set of character images - called \textit{sprites} - and a network that transforms and positions them on a canvas in order to reconstruct input lines. Due to the potential ambiguity in the definition of sprites, we introduce a complementary weak-supervision from line-level transcriptions.

In this section, we first present an overview of our image model and approach (\Cref{sec:overview}). Then, we describe the deep architecture we use (\Cref{sec:archi}). Finally, we discuss our loss and training procedure (\Cref{sec:learning}).

\paragraph{Notations.} We write $a_{1:n}$ the sequence $\{a_1, \dots , a_n\}$, and use bold letters $\mathbf{a}$ for images.
An RGBA image $\mathbf{a}$ corresponds to an RGB image denoted by $\mathbf{a}^c$, alongside an alpha-transparency channel denoted by $\mathbf{a}^\alpha$. We use $\theta$ as a generic notation for network parameters and thus any character indexed by $\theta$, e.g., $a_\theta$,  is a network. 

\subsection{Overview and image model}\label{sec:overview}

\Cref{fig:overview} presents an overview of our pipeline. An input image $\img$ of size $H \times W$ is fed to an encoder network $\encoder$ generating a sequence of $T$ features $\feature_{1:T}$ associated to uniformly-spaced locations $x_{1:T}$ in the image. 
Each feature $\feature_t$ is processed independently by our \textit{Typewriter} module (\Cref{sec:archi}) which outputs an RGBA image $\layer_t$ corresponding to a character. The images $\layer_{1:T}$ are then composited with a canvas image we call $\layer_{T+1}$. This canvas image $\layer_{T+1}$ is a completely opaque image (zero transparency). Its colors are predicted by a Multi-Layer Perceptron (MLP) $\bkg$ which takes as input the features $\feature_{1:T}$ and outputs RGB values $b_{1:T}$. 
All resulting images $\layer_{1:T+1}$ can be seen as ordered image layers and are merged using alpha compositing, as proposed by both~\cite{monnier_unsupervised_2021,smirnov_marionette_2021}. More formally, the reconstructed image $\rec$ can be written:
\begin{equation}
\rec = \sum_{t=1}^{T+1} \Big[\prod_{j<t}(1 - \layera_j)\Big]\layera_t\layerc_t . 
\end{equation}
In practice, we randomize the order of $\layer_{1:T}$ in the compositing operation to reduce overfitting, as advocated by the MarioNette approach~\cite{smirnov_marionette_2021}.
The full system is differentiable and can be trained end-to-end. 

\subsection{Typewriter Module}\label{sec:archi}

We now describe in detail the Typewriter module, which takes as input a feature $\feature$ from the encoder and its position $x$, and outputs an image layer $\layer$, to be composited. An overview of the module is presented in~\Cref{fig:writer}. On a high level, it is similar to the MarioNette architecture~\cite{smirnov_marionette_2021}, but handles blanks (\ie, the generation of a completely transparent image) in a different way and has a more flexible deformation model, similar to the one used in DTI-Sprites~\cite{monnier_unsupervised_2021}. More specifically, the module learns jointly RGBA images called \textit{sprites} corresponding to character images, and networks that use the features $\feature$ to predict a probability for each sprite and a transformation of the sprite. We detail how we obtain the following three elements: the set of $K$ parameterized sprites, the sprites compositing and the transformation model.

\paragraph{Sprite parametrization.}
We model characters as a set of $K$ sprites which are defined using a generator network. More specifically, we learn $K$ latent codes $\scode_{1:K}$ which are used as an input to a generator network $\generator$ in order to generate sprites $\sprite_{1:K}=\generator(\scode_{1:K})$. These sprites are images with a single channel that corresponds to their opacity. Similar to DTI-Sprites~\cite{monnier_unsupervised_2021}, we model a variable number of sprites with an empty (\ie, completely transparent) sprite which we write $\sprite_{K+1}$. In comparison with directly learning sprites in the image space as in DTI-Sprites~\cite{monnier_unsupervised_2021}, we found that using a generator network yields faster and better convergence.

\paragraph{Sprite probabilities and compositing.}
To predict a probability $\proba_k$ for each sprite $\sprite_k$, each latent code $\scode_k$ is associated through a network $p_\theta$ to a probability feature $\pcode_k=p_\theta(\scode_k)$ of the same dimension $D$ as the encoder features ($D=64$ in our experiments). We additionally optimize directly a probability feature $\pcode_{K+1}$ which we associate to the empty sprite. Given a feature $\feature$ predicted by the encoder, we predict the probability $\proba_k$ of each sprite $\sprite_k$ by computing the dot product between the probability features $\pcode_{1:K+1}$ and a learned projection of the feature $\pi_\theta(\feature)$, and applying a $\softmax$ to the result:
\begin{equation}
    \proba_{1:K+1}(f)=\softmax \Big(\lambda z_{1:K+1}^{p} \dotprod \pi_\theta(\feature)^{T}\Big),
\label{eq}
\end{equation}
where $\dotprod$ is the dot product applied to each element of the sequence, $\lambda=1 / \sqrt{D}$ is a scalar temperature hyper-parameter, and the softmax is applied to the resulting vector. 
We use these probabilities to combine the sprites into the weighted average $\sprite=\sum_{k=1}^K \proba_k g_\theta(\scode_k)$. Note that this compositing can be interpreted as attention operation~\cite{vaswani2017attention}: 
\begin{equation}
\sprite = \text{attention}(\bar{Q}, \bar{K}, \bar{V}) = \text{softmax} \left( \frac{\bar{Q}\bar{K}^T}{\sqrt{D}} \right) \bar{V},
\end{equation}
with $\bar{Q}=\pi_\theta(\feature)$, $\bar{K}=p_\theta(\scode_{1:K+1})$, $\bar{V}=g_\theta(\scode_{1:K+1})$, $D$ the dimension of the features, and by convention  $g_\theta(\scode_{K+1})$ is the empty sprite and $p_\theta(\scode_{K+1})=\pcode_{K+1}$.

{We} {actually show that directly optimizing $\pcode_{1:K}$ instead of learning to predict the probability features $\pcode_{1:K}$ from the sprite latent codes $\scode_{1:K}$, similar to {MarioNette}~\cite{smirnov_marionette_2021}, yields similar results.} Note that we learn a probability code $\pcode_{K+1}$ to compute the probability of empty sprites instead of having a separate mechanism as in {MarioNette}~\cite{smirnov_marionette_2021} because it is critical for our supervised loss (see~\cref{sec:sup}). 

\paragraph{Positioning and coloring.}
The final step of our module is to position the selected sprite in a canvas of size $H \times W$ and to adapt its color. We implement this operation as a sequence of a spatial transformer~\cite{jaderberg_spatial_2015} and a color transformation, similar to {DTI-Sprites}~\cite{monnier_unsupervised_2021}. {More specifically, the feature $\feature$ is given as input to a network $t_\theta$ that predicts three parameters for the color of the sprite and three parameters for isotropic scaling and 2D-translation that are used by a spatial transformer \cite{jaderberg_spatial_2015} to deform $\sprite$. 
Finally, using the location $x$ associated to the feature $\feature$, we paste the deformed colored sprite onto a background canvas of size $H \times W$ at position $x$ to obtain a reconstructed RGBA image layer $\layer$. Positioning the sprites with respect to the position of the associated local features helps us obtain results co-variant to translations of the text lines and independent of the line size. To produce the background canvas, the features $f_{1:T}$ are first each passed through a shared MLP $\bkg$, to produce background colors $b_{1:T}$. We then use bi-linear interpolation to upscale these $T$ colors to fit the size of the input image. Details on the parametrization of the transformation networks are presented in the supplementary material.} 

\begin{figure*}[!ht]
  \begin{subfigure}{0.48\linewidth}
  \centering
    \includegraphics[width=1.0\textwidth]{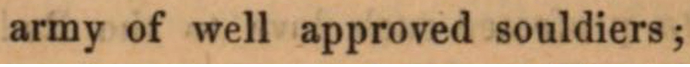}\\
    \includegraphics[width=1.0\textwidth]{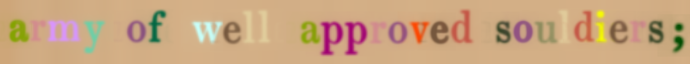} \\
    \includegraphics[width=1.0\textwidth]{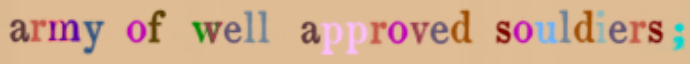} \\    \caption{Input, supervised and unsupervised semantic segmenation.}
    \label{fig:google_results}
  \end{subfigure}
  \begin{subfigure}{0.24\linewidth}
  \centering
    \includegraphics[width=1.0\textwidth]{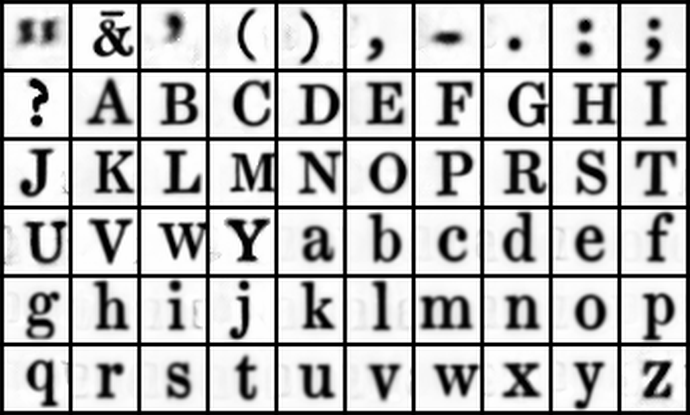}
    \caption{Supervised Sprites.}
    \label{fig:google_supervised}
  \end{subfigure}
  \begin{subfigure}{0.24\linewidth}
  \centering
    \includegraphics[width=1.0\textwidth]{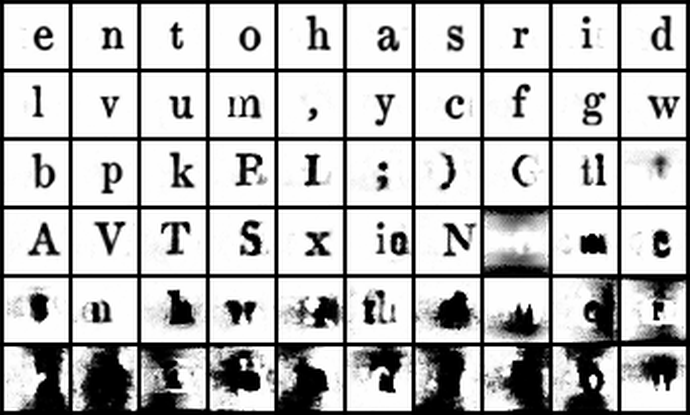}
    \caption{Unsupervised Sprites.}
    \label{fig:google_unsupervised}
  \end{subfigure}
  \caption{\textbf{Results on a printed document (Google1000).} In both the supervised and unsupervised setting our method produces meaningful sprites and accurate reconstructions (\ref{fig:google_results}). We show the 60 most used sprites in alphabetic ordering in the supervised setting (\ref{fig:google_supervised}) and ordered by frequency in the unsupervised one (\ref{fig:google_unsupervised}). See text for details and the supplementary material for more reconstructions. 
  }
  \label{fig:qualitative-google}
  \vspace{-0.6em}
\end{figure*}

\subsection{Losses and training details}\label{sec:learning}

Our system is designed in an analysis-by-synthesis spirit, and thus relies mainly on a reconstruction loss. This reconstruction loss can be complemented by a loss on the selected sprites in the supervised setting where each text line is paired with a transcription. In the following, we define these losses for a single text line image and its transcription, using the notations of the previous section. 
\paragraph{Reconstruction loss.} Our core loss is a simple mean square error between the input image $\img$ and its reconstruction $\rec$ predicted by our system as described in~\cref{sec:overview}:

\begin{equation}
     \lrec(\img,\rec )=\| \img - \rec \|^2
\label{eq:unsup}.
\end{equation}
In the unsupervised setting, we use this loss alone without any additional regularization.

\paragraph{Weakly supervised loss.}\label{sec:sup}
The intrinsic ambiguity of the sprite decomposition problem may result in sprites that do not correspond to individual characters.
Using line-level annotation is an easy way to resolve this ambiguity. We find that simply adding the classical CTC loss~\cite{ctc} computed on the sprite probabilities to our reconstruction loss is enough to learn sprites that exactly correspond to characters. More specifically, we chose the number of sprites as the number of different characters and associate arbitrarily each sprite to a character and the empty sprite to the separator token of the CTC. Then given the one-hot line-level annotation $\gt$ and the predicted sprite probabilities $\pred = (\proba_{1:K+1}(f_{1}), ..., \proba_{1:K+1}(f_T))$, we optimize our system's parameters by minimizing:
\begin{equation}
    \lsup(\img,\gt ,\rec,\pred) =  \lrec(\img,\rec )  + \lambdactc \lctc(\gt ,\pred)
\label{eq:sup}
\end{equation}
where $\lambdactc$ is a hyper-parameter and $\lctc(\gt ,\pred)$ is the CTC loss computed between the ground-truth $\gt$ and the predicted probabilities $\pred$. In our experiments we have used $\lambdactc=0.1$ for printed text and $\lambdactc=0.01$ for handwritten text.

\paragraph{Implementation and training details.}
{We train on the Google1000~\cite{vincent_google_2007} and Fontenay~\cite{pal} datasets with lines of height $H=64$ and on the Copiale dataset~\cite{knight_copiale_2011} with $H=92$. The generated sprites $\sprite_{1:K}$ are of size $\nicefrac{H}{2} \times \nicefrac{H}{2}$. In the supervised setting, we use as many sprites as there are characters, and in the unsupervised we set $K=60$ for Google1000 and $K=120$ for the Copiale cipher. In the supervised case we train for 100 epochs on Google1000 and for 500 epochs on Copiale with a batch size of 16, and we select the model that performs best on the validation set for evaluation. In the unsupervised setting we use line crops of width $W=2H$ and train for 1000 epochs on Google1000 and for 5000 on the Copiale cipher with a batch size of 32 and use the final model. The number of epochs is much higher in the unsupervised case than in the supervised case because the network sees only a small crop of each line at each epoch, but each epoch is much faster to perform.}

Our encoder network is a ResNet-32-CIFAR10 \cite{he2016deep}, that is truncated after layer 3 with a Gaussian feature pooling described in supplementary material. For our unsupervised experiments, we use as generator $\generator$ the U-Net architecture of Deformable Sprites~\cite{vicky-unet} which converged quickly, and for our supervised experiments a 2-layer MLP similar to MarioNette~\cite{smirnov_marionette_2021} which produces sprites of higher quality. The networks $\pi_\theta$ and $\proba_\theta$ are a single linear layers followed by layer-normalization. We use the AdamW~\cite{adamw} optimizer with a learning rate of $10^{-4}$ and apply a weight-decay of $10^{-6}$ to the encoder parameters. At inference we select the sprites with the highest probabilities instead of using a softmax.

\begin{figure*}[!ht]
  \begin{subfigure}{0.28\linewidth}
    \centering
    \includegraphics[width=1.0\textwidth, height=0.03\textheight]{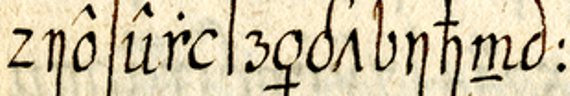}\\
    \includegraphics[width=1.0\textwidth, height=0.03\textheight]{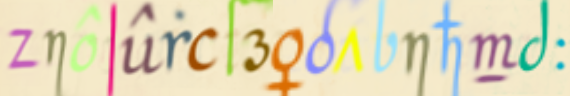} \\
    \includegraphics[width=1.0\textwidth, height=0.03\textheight]{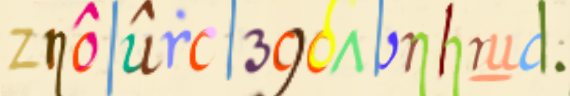} \\
    \caption{Input, supervised and unsupervised.}
    \label{fig:copiale_results}
  \end{subfigure}
  \hfill
  \begin{subfigure}{0.35\linewidth}
  \centering
    \includegraphics[width=1.0\textwidth]{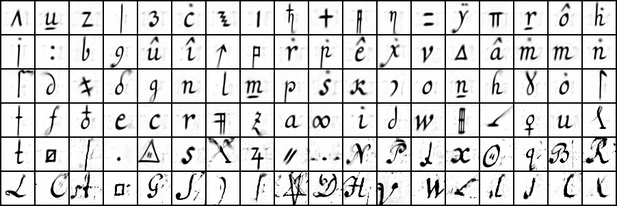}
    \caption{Supervised Sprites.}
    \label{fig:copiale_supervised}
  \end{subfigure}
  \hfill
  \begin{subfigure}{0.35\linewidth}
  \centering
    \includegraphics[width=1.0\textwidth]{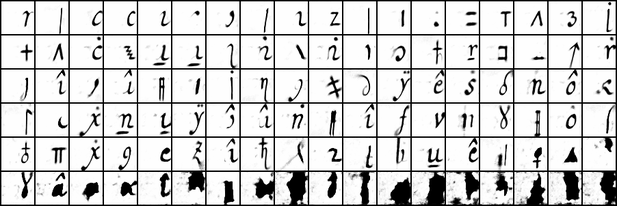}
    \caption{Unsupervised Sprites.}
    \label{fig:copiale_unsupervised}
  \end{subfigure}
  \caption{{\textbf{Results on the Copiale cipher~\cite{knight_copiale_2011}.} Despite the high number of characters and their variability, our method learns meaningful sprites and performs accurate reconstructions in both settings (\ref{fig:copiale_results}). We show the 108 most used sprites sorted by frequency in 
 the supervised (\ref{fig:copiale_supervised}) and the unsupervised (\ref{fig:copiale_unsupervised}) settings.}}
  \label{fig:qualitative-copiale}
  \vspace{-0.3em}

\end{figure*}

\section{Experiments}\label{sec:experiments}
\subsection{Datasets and metrics}\label{datasets}
\paragraph{Datasets.} We experiment with three datasets with different characteristics: Google1000~\cite{vincent_google_2007}, the Copiale cipher~\cite{knight_copiale_2011} and Fontenay manuscripts~\cite{pal,dh2022}:

\begin{itemize}[itemsep=0em, wide, labelindent=0pt, topsep=.5em]%
\item \textit{Google1000.} The Google1000 dataset contains scanned historical printed books, arranged into Volumes~\cite{vincent_google_2007}. We use the English Volume 0002 which we process with the preprocessing code of~\cite{gupta_learning_2018}, using 317 out of 374 pages and train-val-test set with 5097-567-630 lines respectively. This leads to a total number of 83 distinct annotated characters. 
Although supervised printed font recognition is largely considered a solved problem, and the annotation for this dataset are actually the result of OCR, this document is still challenging for an analysis-by-synthesis approach, containing artifacts such as ink bleed, age degradation, as well as variance in illumination and geometric deformations.
\item\textit{Copiale cipher.} The Copiale cipher is an oculist German text dating back to a 18th century secret society~\cite{knight_copiale_2011}. Opposite to Baro et al.~\cite{baro_generic_2019} which uses a binarized version of the dataset, we train our model on the original text-line images, which we segmented using docExtractor~\cite{monnier_docextractor_2020} and manually assigned to the annotations, respecting the train-val-test split of Baro et al.~\cite{baro_generic_2019} with 711-156-908 lines each. The total number of distinct annotated characters is 112.
This dataset is more challenging than printed text because because it is handwritten, which introduces some variability in the character shapes, and because of the large number of characters.

\item \textit{Fontenay manuscripts.} The Fontenay dataset contains digitized charters that originate from the Cistercian abbey of Fontenay in Burgundy (France)~\cite{pal,dh2022} and were created during the 12th and early 13th century. Each document has been digitized and each line has been manually segmented and transcribed. For our experiments, we selected a subset of 14 different documents sharing a similar script which falls into the family of praegothica scripts. These correspond to 163 lines, using 47 distinct characters. While they were carefully written and preserved, these documents are still very challenging (\Cref{fig:paleo_qual}). They exhibit degradation, clear intra-document letter shape variations, and letters can overlap or be joined by ligature marks. Moreover, each document represents only a small amount of data, e.g., the ones used in~\Cref{fig:paleo_qual} contain between 8 and 25 lines.
\end{itemize}

\paragraph{Metrics.}
\vspace{-1em}
Our goal is to capture the shape of all characters and position them precisely in each text line. Such fine-grained annotation is however not available in existing datasets. Instead, to provide a quantitative evaluation of our models, we report L2 reconstruction error ('Rec.' in the tables) and Character Error Rate (CER). CER is the standard metric for Optical Character Recognition (OCR). Given ground-truth and predicted sequences of characters, $\sigma$ and $\hat{\sigma}$, it is defined as the minimum number of substitutions $S$, deletions $D$, and insertions $I$ of characters necessary to match the predicted sequence $\hat{\sigma}$ to the ground truth sequence $\sigma$, normalized by the size of the ground truth sequence $|\sigma|$:

\begin{equation}
    {CER(\sigma, \hat{\sigma}) = \frac{S + D + I}{|\sigma|}.}
\end{equation}

For simplicity, we ignore spaces. Predictions are obtained by selecting at every position the character associated to the most likely sprite. {In the supervised setting, the association between sprites and characters is fixed at the beginning of training. In the unsupervised setting, we associate every sprite to a single character using a simple assignment strategy described in supplementary material. More complex assignments, for example associating sprite bi-grams to individual characters, or even incorporating their relative positions, could be considered for a recognition performance boost. However, since OCR is not our main goal but simply a proxy measure, this falls out of the scope of our work.}

\subsection{Qualitative results}\label{sec:qual_results}
Examples of semantic segmentation and sprites in the supervised and unsupervised setting on Google1000 and Copiale are shown in~\Cref{fig:qualitative-google,fig:qualitative-copiale} respectively. 

In the unsupervised setting, several sprites (\Cref{fig:google_unsupervised,fig:copiale_unsupervised})
can be used to reconstruct a single character. For example, the 'n' and 'm' sprites are joined in order to better reconstruct 'm' in Google1000. 
To account for appearance variation, multiple sprites are learned to reconstruct the most frequent character, e.g. 'e' for Google and 'c' in Copiale. This effects are even stronger in the handwritten Copiale dataset, where generic sub-character strokes are learned and used together to better model characters' variations.
In both datasets, a portion of the least used sprites are not well optimized, do not correspond to characters, and are not used by the network. 
These behaviors are expected in a completely unsupervised setting, because of the highly unbalanced statistics of the character frequencies and ambiguity of the reconstruction: without additional supervision, there is a clear benefit for the network to model well variations of common characters, and to approximate or discard rare ones. This is a core limitation of existing unsupervised image decomposition approaches, and a motivation for the introduction of our weakly supervised setting.

In the (weakly) supervised setting, the sprites (\Cref{fig:google_supervised,fig:copiale_supervised}) closely correspond to the characters, with the exception of very rare characters like the capital 'Z' character for Google1000 (as can be seen in supplementary material), while reconstruction is of very high quality and each character is reconstructed with the expected sprite.

\subsection{Quantitative results} 
Our quantitative results and ablations for Google1000 and Copiale are reported in \Cref{tab:google,tab:copiale} respectively. 

For Google1000, the CER in the supervised setting is close to perfect, while it is $7.7\%$ for the unsupervised setting.
To provide baselines for these performances, we trained on our data (i) ScrabbleGAN~\cite{fogel_scrabblegan_2020}, a supervised method with a standard recognizer and an additional generator module, (ii) FontAdaptor~\cite{zhang_adaptive_2020}, a recent 1-shot method that learns to match single character examplars to text lines, and (iii)  an adaptation of the unsupervised DTI-Sprites~\cite{monnier_deep_2020} to text lines 
which we detail in {the} supplementary material (we also show in the supplementary material that vanilla MarioNette~\cite{smirnov_marionette_2021} provides clearly worse results.). 
Our unsupervised approach performs clearly better than our adaptation of DTI-Sprites and is almost on par with the 1-shot FontAdaptor, while our weakly supervised approach is almost on par with ScrabbleGAN. Our adaptation of DTI-Sprites is better reconstruct images, but the learned sprites are much less meaningful, as shown by the poor CER performance. Interestingly, reconstruction is much better when using supervision, which hints that a better optimization scheme might help improving unsupervised performances. We also evaluated the effect of varying the number of sprites $K$ in the unsupervised setting. For half the number of sprites ($K=30$) of compared to our baseline ($K=60$) we observe a significant performance drop to $15.01 \pm 5.9\%$ CER, while increasing the number of sprites to $K=120$ and even $K=240$ only slightly boosts performances with $6.9 \pm 0.6\%$ and $5.7 \pm 1.6\%$ CER.

 
On the Copiale dataset, we compare our results with HTRbyMatching~\cite{HTRbyMatching}, a few-shot approach developed specifically for cipher recognition, using the same train/val/test splits. HTRbyMatching was evaluated on a wide range of few-shot scenarios, ranging from a scenario similar to FontAdaptor where a single exemplar is available for every character, to one where 5 exemplars are available for each character together with 5 completely annotated pages. Reported results are only for confident character predictions with different confidence thresholds, but summing the error rate of the predicted symbols and the percentage of non-annotated symbols, one can estimate the CER to vary between 10\% and 47\% depending on the scenario. This is consistent with the quantitative results we obtain with our approach, which are much better in the supervised setting ({$4.2\%$}) and worse in the completely unsupervised one ($52.6\%$). The low performance of the unsupervised approach is consistent with the qualitative results: given that many characters are reconstructed are reconstructed by sub-character sprites, one would have to associate sprite bi-grams to characters in order to obtain good CER performances. Interestingly, the reconstruction error is similar in the supervised and unsupervised setting, hinting that for this specific dataset, optimizing the reconstruction quality might not be enough to obtain relevant decomposition without additional priors. These results enables to quantify and analyze a limitation of unsupervised image decomposition approaches on a more challenging dataset.

Note that the goal of our approach is not to boost CER performances - which in any case would be meaningless on Google1000 where the ground truth is already the result of an OCR model - but instead to learn character models and image decomposition, and all these comparisons should be considered as sanity checks. Designing post-processing to improve CER is possible, for example we tested a simple post-processing associating new sprites to the most frequent bi-grams and tri-grams, that leads to an improved CER for Copiale of 29.9\%. However, we think it is more interesting to see this metric as a tool to evaluate the raw output of unsupervised decomposition models.


\begin{table}[t]
\small
\centering
 \begin{tabular}{@{}lccc@{}}
 \toprule
Method & Type & Rec. $\times\;10^{3}$ & CER  \\
\midrule
DTI-Sprites~\cite{monnier_unsupervised_2021} & unsup. & {2.54
} & 18.4 \%\\
FontAdaptor~\cite{zhang_adaptive_2020} & 1-shot & - & 6.7 \%\\
ScrabbleGAN~\cite{fogel_scrabblegan_2020} & sup. & - & 0.6 \%\\
\midrule
Learnable Typewriter & sup. & $3.5 \pm 0.1$ & {$0.85 \pm 0.03\%$} \\
\hspace{1em} w\textbackslash o shared $z_{k}$ & sup. & $3.3 \pm 0.1$ & $0.89\pm 0.06\%$\\
\hspace{1em} w\textbackslash o  $p_{\theta}$ & sup. & $3.5 \pm 0.1$ & $0.99 \pm 0.05\%$\\
\hspace{1em} w\textbackslash o  $g_{\theta}$ & sup. & $3.4 \pm 0.1$  & $0.88 \pm 0.04\%$\\



Learnable Typewriter & unsup. & $7.1 \pm 0.4$ & $7.7 \pm 0.6\%$\\
\hspace{1em} w\textbackslash o shared $z_{k}$ & unsup. & $7.4 \pm 0.4$ & $8.0 \pm 0.2\%$\\
\hspace{1em} w\textbackslash o  $p_{\theta}$ & unsup. & $7.0 \pm 0.3$ & $7.7 \pm 2.0\%$\\
\hspace{1em} w\textbackslash o  $g_{\theta}$ & unsup. & $10.5 \pm 0.7$  & $27.0 \pm 2.2\%$\\
\bottomrule
\end{tabular}



\caption{\textbf{Quantitative results and ablation on Google1000~\cite{vincent_google_2007}.} We report CER and L2 reconstruction error for different approaches. For our method, we report average over 5 runs and standard deviation. 
}
\vspace{-0.4em}

\label{tab:google}
\end{table}

In particular, we performed on both datasets an ablation of the architecture 
to better understand which design choices are critical. Interestingly, our results show that both in the supervised and the unsupervised setting, not sharing the latent codes $z_{k}$ between the generation network and the sprite selection and even completely removing the probability network $p_{\theta}$ has limited influence on the performance, and these design choices of MarioNette~\cite{smirnov_marionette_2021} are not critical. 
Conversely, removing $g_\theta$ and directly learning protototypes as network parameters similar to DTI-Sprites~\cite{monnier_unsupervised_2021} has little impact in the supervised case but leads to significant drops in performance. A more detailed analysis of training curves reveals that training is slower and overfits. While it might be possible to fix this issue by adapting the learning scheme for the prototypes, it shows it is easier to learn the prototypes through a generator network than optimizing them directly.

\begin{table}[t]
\small
\centering
 \begin{tabular}{@{}lccc@{}}
 \toprule
Method & Type & Rec. $\times 10^{2}$ & CER  \\ \midrule
HTRbyMatching~\cite{HTRbyMatching} & few-shot & - & $10-47\% ^*$\\
\midrule
Learnable Typewriter & sup. & $1.81 \pm 0.01$ & $4.2 \pm 0.3\%$ \\ 
\hspace{1em} w\textbackslash o shared $z_{k}$ & sup. & $1.79 \pm 0.01$ & $4.0 \pm 0.1\%$ \\
\hspace{1em} w\textbackslash o  $p_{\theta}$ & sup.  & $1.77 \pm 0.02$ & $4.7 \pm 0.1\%$ \\
\hspace{1em} w\textbackslash o  $g_{\theta}$ & sup. & $1.96 \pm 0.07$  & $4.2 \pm 0.2\%$\\

Learnable Typewriter & unsup. & $1.93 \pm 0.02$ & {$52.6 \pm 1.7 \%$} \\
\hspace{1em} w\textbackslash o shared $z_{k}$ & unsup. & $1.89 \pm 0.02$ & $47.6 \pm 2.8\%$ \\
\hspace{1em}  w\textbackslash o  $p_{\theta}$ & unsup. & $1.81 \pm 0.06$ & $51.9 \pm 2.0\%$ \\
\hspace{1em} w\textbackslash o  $g_{\theta}$ & unsup. & $3.99 \pm 0.14$  & $80.6 \pm 0.9\%$\\

\bottomrule
\end{tabular}
\caption{\textbf{Quantitative results on Copiale~\cite{knight_copiale_2011}.} {We report CER and reconstruction error for a baselines and our method. For our method, we report average over 5 runs and standard deviation. $^*$See text for details. }}
\label{tab:copiale}
\vspace{-0.3em}
\end{table}

\subsection{Application to paleography}
 
To test our approach in a more challenging case and demonstrate its potential for paleographic analysis, we applied it on a collection of 14 historical charters from the Fontenay abbey~\cite{pal,dh2022}. While they all use similar scripts from the Praegothica type, they also exhibit clear variations. One of the goals of a {paleographic} analysis would be to identify and characterize these variations. {In line with the seminal System for Paleographic Inspections \cite{SPI, SPI-2}, we focus on the variations in the shape of letters, which are quite challenging to describe with natural language.} One solution would be to choose a specific example for each letter in each document or to have a paleographer manually draw of a 'typical' one. However, this is very time consuming and might reflect priors or bias from the paleographer in addition to the actual variations. Instead, we propose to fine-tune our Learnable Typewriter approach on each document and visualize the sprites associated to each character and each document. Because of the difficulty of the dataset, we focus on the results of our supervised setting.


\begin{figure}[t]
  \centering
  \begin{subfigure}{1.0\linewidth}
  \centering
    \includegraphics[height=1.35cm,cfbox=yellow 0.1pt 0pt]{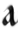}\hspace{1mm}
    \includegraphics[height=1.35cm]{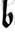}\hspace{1mm}
    \includegraphics[height=1.35cm]{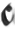}\hspace{1mm}
    \includegraphics[height=1.35cm]{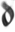}\hspace{1mm}
    \includegraphics[height=1.35cm]{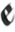}\hspace{1mm}
    \includegraphics[height=1.35cm]{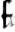}\hspace{1mm}
    \includegraphics[height=1.35cm,cfbox=magenta 0.1pt 0pt]{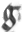}\hspace{1mm}
    \includegraphics[height=1.35cm]{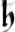}\hspace{1mm}
\end{subfigure}
  \begin{subfigure}{1.0\linewidth}
  \centering
    \includegraphics[height=1.35cm,cfbox=yellow 0.1pt 0pt]{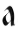}\hspace{1mm}
    \includegraphics[height=1.35cm]{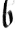}\hspace{1mm}
    \includegraphics[height=1.35cm]{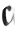}\hspace{1mm}
    \includegraphics[height=1.35cm]{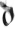}\hspace{1mm}
    \includegraphics[height=1.35cm]{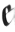}\hspace{1mm}
    \includegraphics[height=1.35cm]{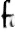}\hspace{1mm}
    \includegraphics[height=1.35cm,cfbox=magenta 0.1pt 0pt]{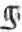}\hspace{1mm}
    \includegraphics[height=1.35cm]{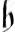}\hspace{1mm}
\end{subfigure}
   \begin{subfigure}{1.0\linewidth}
  \centering
    \includegraphics[height=1.35cm,cfbox=yellow 0.1pt 0pt]{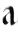}\hspace{1mm}
    \includegraphics[height=1.35cm]{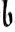}\hspace{1mm}
    \includegraphics[height=1.35cm]{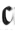}\hspace{1mm}
    \includegraphics[height=1.35cm]{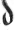}\hspace{1mm}
    \includegraphics[height=1.35cm]{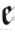}\hspace{1mm}
    \includegraphics[height=1.35cm]{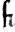}\hspace{1mm}
    \includegraphics[height=1.35cm,cfbox=magenta 0.1pt 0pt]{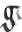}\hspace{1mm}
    \includegraphics[height=1.35cm]{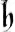}\hspace{1mm}
\end{subfigure}
  \begin{subfigure}{1.0\linewidth}
  \centering
    \includegraphics[height=1.35cm,cfbox=yellow 0.1pt 0pt]{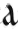}\hspace{1mm}
    \includegraphics[height=1.35cm]{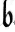}\hspace{1mm}
    \includegraphics[height=1.35cm]{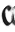}\hspace{1mm}
    \includegraphics[height=1.35cm,cfbox=blue 0.1pt 0pt]{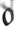}\hspace{1mm}
    \includegraphics[height=1.35cm]{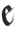}\hspace{1mm}
    \includegraphics[height=1.35cm]{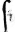}\hspace{1mm}
    \includegraphics[height=1.35cm,cfbox=magenta 0.1pt 0pt]{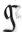}\hspace{1mm}
    \includegraphics[height=1.35cm]{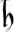}\hspace{1mm}
  \end{subfigure}
  \begin{subfigure}{1.0\linewidth}
  \centering
    \includegraphics[height=1.35cm,cfbox=yellow 0.1pt 0pt]{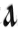}\hspace{1mm}
    \includegraphics[height=1.35cm]{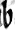}\hspace{1mm}
    \includegraphics[height=1.35cm]{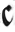}\hspace{1mm}
    \includegraphics[height=1.35cm]{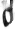}\hspace{1mm}
    \includegraphics[height=1.35cm,cfbox=green 0.1pt 0pt]{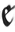}\hspace{1mm}
    \includegraphics[height=1.35cm]{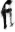}\hspace{1mm}
    \includegraphics[height=1.35cm,cfbox=magenta 0.1pt 0pt]{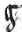}\hspace{1mm}
    \includegraphics[height=1.35cm]{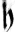}\hspace{1mm}
    \end{subfigure}
  \caption{\textbf{Sprites learned for similar documents in Praegothica script.} Each line corresponds to a different document. Looking at any column, one can notice the small differences that characterise the handwriting in each document. Colored boxed correspond to cases analysed in more details in~\Cref{fig:paleo_qual}.}
  \label{fig:paleo_main}
  \vspace{-0.6em}
\end{figure}

\begin{figure}
\begin{subfigure}{1.0\linewidth}
  \centering
\includegraphics[height=1.04cm,cfbox=yellow 0.1pt 0pt]{paper/paleo/pal/gray_s_6_doc_0.png}
    \includegraphics[height=1.04cm]{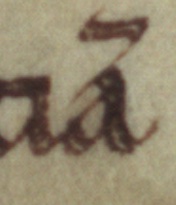}
    \includegraphics[height=1.04cm,cfbox=yellow 0.1pt 0pt]{paper/paleo/pal/gray_s_6_doc_1.png}
    \includegraphics[height=1.04cm]{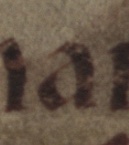}
    \includegraphics[height=1.04cm,cfbox=yellow 0.1pt 0pt]{paper/paleo/pal/gray_s_6_doc_2.png}
    \includegraphics[height=1.04cm]{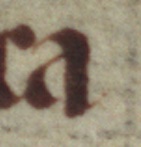}
    \includegraphics[height=1.04cm,cfbox=yellow 0.1pt 0pt]{paper/paleo/pal/gray_s_6_doc_3.png}
    \includegraphics[height=1.04cm]{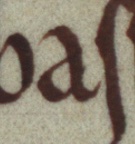}
    \includegraphics[height=1.04cm,cfbox=yellow 0.1pt 0pt]{paper/paleo/pal/gray_s_6_doc_4.png}
    \includegraphics[height=1.04cm]{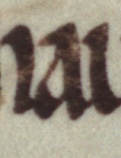}\\[2mm]
    \includegraphics[height=1.01cm,cfbox=magenta 0.1pt 0pt]{paper/paleo/pal/gray_s_12_doc_0.png}
    \includegraphics[height=1.01cm]{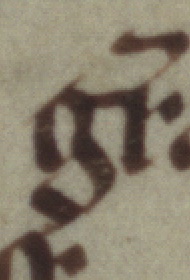}
    \includegraphics[height=1.01cm,cfbox=magenta 0.1pt 0pt]{paper/paleo/pal/gray_s_12_doc_1.png}
    \includegraphics[height=1.01cm]{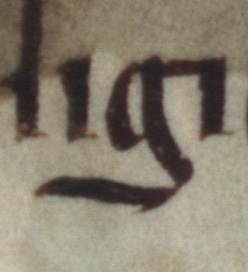}
    \includegraphics[height=1.01cm,cfbox=magenta 0.1pt 0pt]{paper/paleo/pal/gray_s_12_doc_2.png}
    \includegraphics[height=1.01cm]{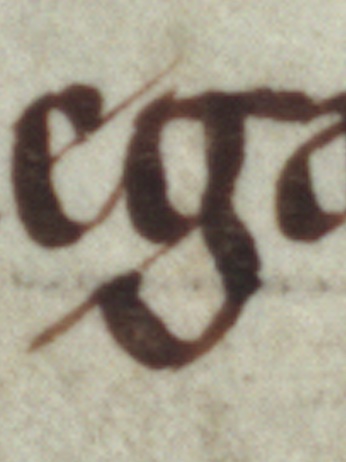}
    \includegraphics[height=1.01cm,cfbox=magenta 0.1pt 0pt]{paper/paleo/pal/gray_s_12_doc_3.png}
    \includegraphics[height=1.01cm]{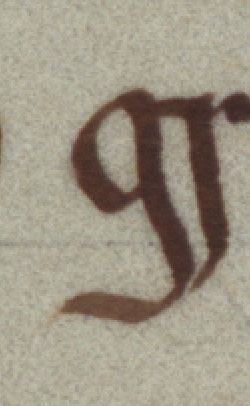}
    \includegraphics[height=1.01cm,cfbox=magenta 0.1pt 0pt]{paper/paleo/pal/gray_s_12_doc_4.png}
    \includegraphics[height=1.01cm]{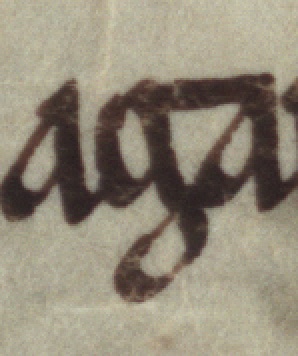}
    \caption{'a' and 'g' sprite for each document and associated example of the character. Note how the variations of the descending part of the 'g' sprites closely match the variations observed in the documents. Also note the subtle variations of the 'a' which are clear in the sprites but would be hard to notice and describe from the original images for a non-expert.}  
  \label{fig:paleo_ag}
  \end{subfigure}
  \begin{subfigure}{1.0\linewidth}
  \centering
    \includegraphics[height=1.25cm,cfbox=green 0.1pt 0pt]{paper/paleo/pal/gray_s_10_doc_4.png}\hspace{1mm}
    \includegraphics[height=1.25cm]{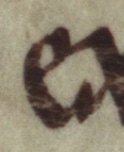}
    \includegraphics[height=1.25cm]{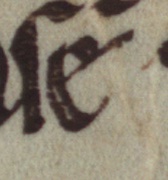}
    \includegraphics[height=1.25cm]{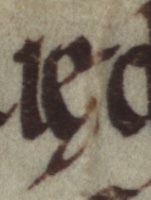}
    \includegraphics[height=1.25cm]{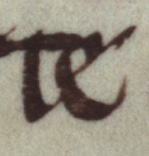}
    \includegraphics[height=1.25cm]{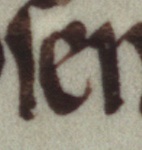}
    \includegraphics[height=1.25cm]{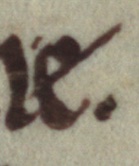}
      \caption{The appearance variations of individual instances associated to the 'e' character in the document are accurately visually summarized by the sprite.}
        \label{fig:paleo_e}
  \end{subfigure}
  \begin{subfigure}{1.0\linewidth}
  \centering
    \hspace{-0.5em}
    \includegraphics[height=1.25cm,cfbox=blue 0.1pt 0pt]{paper/paleo/pal/gray_s_9_doc_3.png}
    \hspace{-0.4em}
    \includegraphics[height=1.25cm]{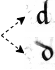}
    \includegraphics[height=1.25cm]{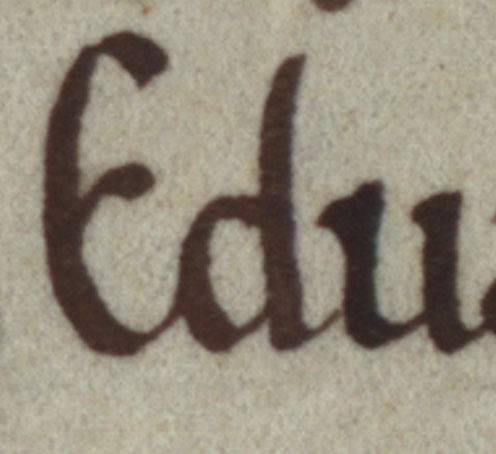}
    \includegraphics[height=1.25cm]{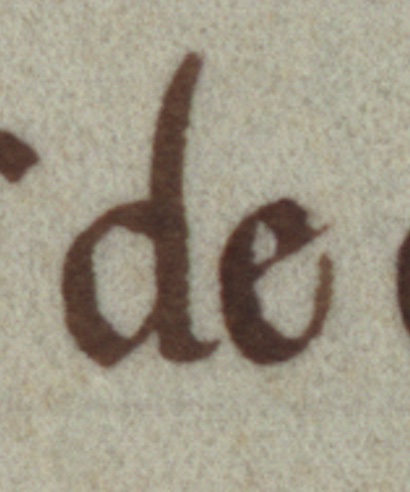}
    \includegraphics[height=1.25cm]{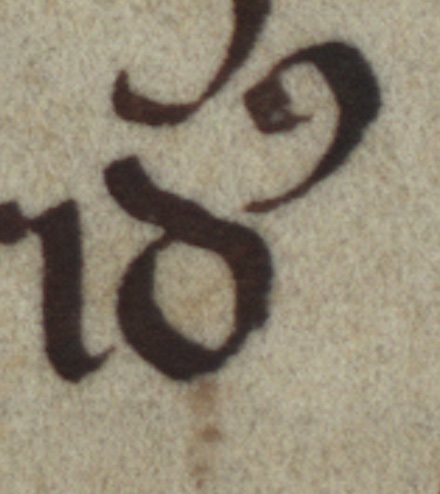}
    \includegraphics[height=1.25cm]{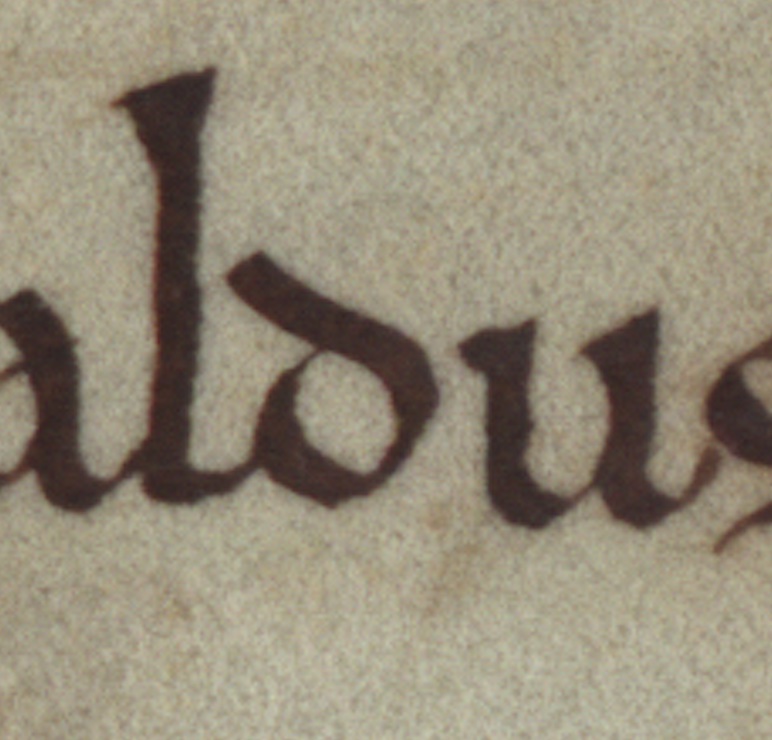}
    \includegraphics[height=1.25cm]{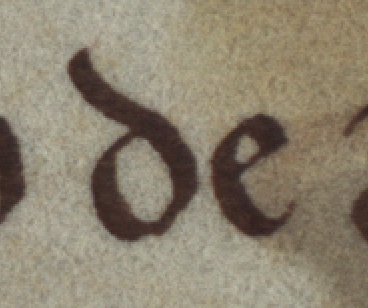}
    \caption{The double appearance of the ascending line of the 'd' sprite shown on the left is related to the co-existence of two different kinds of 'd' in the document, as shown in the examples on the right. {
    We can actually learn both appearances of 'd', shown after the arrows, if we model every character using two sprites.}}
  \label{fig:paleo_d}
  \end{subfigure}
    \caption{The sprites summarize the key attributes of a character in each specific document, averaging its variations. Note the complexity of the documents: characters can overlap or be connected ligature, the parchment is often stained, and there are important intra-document character variations.}
  \label{fig:paleo_qual}
  \vspace{-0.3em}
\end{figure}

\Cref{fig:paleo_main} visualizes the sprites obtained for five different documents from the characters 'a' to 'h' and~\Cref{fig:paleo_qual} highlights different aspects of the results. \Cref{fig:paleo_ag} emphasizes the fact that the differences in the learned sprites correspond to actual variations in the different documents, whether subtle, such as for the 'a' sprite, or clearer, such as for the descending part of the 'g' sprite. \Cref{fig:paleo_e} shows how a sharp sprite can be learned for the character 'e', summarizing accurately its shape despite variations in the different occurrences. Finally, \Cref{fig:paleo_d} shows the case of a document in which two types of 'd' co-exist. In this case, the learned sprite, shown on the left, reassembles an average of the two, with both versions of the ascending parts visible with intermediate transparency. Such a limitation could be overcome by learning several sprites per character. We thus experimented with learning two per character, simply by summing their probabilities when optimizing the CTC-loss. We find that when different appearances of the same letter exist, the two sprites learn two different appearances, and we show the example of the two different learned 'd' sprites on the right of the original one.

Our approach could benefit paleographic analysis in more ways than simply analyzing the characters shapes. Indeed, our model also gives access to the position and scale variation for each letter. This would enable a quantitative analysis of more global appearance factors of the text, related to the space between letters or their respective size variations. Because they would be tremendously tedious to annotate, such variations have rarely been quantified, and their analysis could open new research topics, for example the study of the handwriting evolution of a single writer copying a book across several months.
Another natural application of our approach is font or writer classification, which could be achieved either using a single model to compare errors statistics for the different letters or relative positions of bi-grams, or by training different models for different fonts or writers. The main advantage compared to most existing approaches would be the high interpretability of the predictions, which a user could easily validate.

\section{Conclusion}

We have presented a document-specific generative approach to document analysis. Inspired by deep unsupervised multi-object segmentation methods, we extended them to accurately model standard printed documents as well {as} much more complex ones, such as a handwritten ciphered manuscript or ancient charters. We outlined that a completely unsupervised approach suffers from the ambiguity of the decomposition problem and imbalance characters distributions. We thus extended these approaches using weak supervision to obtain high-quality results. 
Finally, we demonstrated the potential of our Learnable Typewriter approach for a novel application: paleographic analysis.


\section*{Acknowledgments}
We would like to thank Malamatenia Vlachou and Dominique Stutzmann for sharing ideas, insights and data for applying our method in paleography; Vickie Ye and Dmitriy Smirnov for useful insights and discussions; Romain Loiseau, Mathis Petrovich, Elliot Vincent, Sonat Baltacı for manuscript feedback and constructive insights. This work was partly supported by the European Research Council (ERC project DISCOVER, number 101076028), ANR project EnHerit ANR-17-CE23-0008, ANR project VHS ANR-21-CE38-0008 and HPC resources from GENCI-IDRIS (2022-AD011012780R1, AD011012905).


{
\small
\bibliographystyle{ieee_fullname}
\bibliography{main}

\begin{thebibliography}{10}\itemsep=-1pt

\bibitem{SPI}
Fabio Aiolli, Maria Simi, Diego Sona, Alessandro Sperduti, Antonina Starita,
  and Gabriele Zaccagnini.
\newblock Spi: a system for palaeographic inspections.
\newblock {\em AIIA Notizie http://www. dsi. unifi. it/AIIA}, 4:34--38, 1999.

\bibitem{baird1999model}
Henry~S Baird.
\newblock Model-directed document image analysis.
\newblock In {\em Proceedings of the Symposium on Document Image Understanding
  Technology}, volume~1, page~3, 1999.

\bibitem{baro_generic_2019}
Arnau Bar{\'o}, Jialuo Chen, Alicia Forn{\'e}s, and Be{\'a}ta Megyesi.
\newblock Towards a {{Generic Unsupervised Method}} for {{Transcription}} of
  {{Encoded Manuscripts}}.
\newblock In {\em Proceedings of the 3rd {{International Conference}} on
  {{Digital Access}} to {{Textual Cultural Heritage}}}, pages 73--78, {Brussels
  Belgium}, May 2019. {ACM}.

\bibitem{berg-kirkpatrick_unsupervised_2013}
Taylor {Berg-Kirkpatrick}, Greg Durrett, and Dan Klein.
\newblock Unsupervised {{Transcription}} of {{Historical Documents}}.
\newblock In {\em Proceedings of the 51st {{Annual Meeting}} of the
  {{Association}} for {{Computational Linguistics}}}, volume~1, pages 207--217,
  2013.

\bibitem{bluche2017gated}
Th{\'e}odore Bluche and Ronaldo Messina.
\newblock Gated convolutional recurrent neural networks for multilingual
  handwriting recognition.
\newblock In {\em 2017 14th IAPR international conference on document analysis
  and recognition (ICDAR)}, volume~1, pages 646--651. IEEE, 2017.

\bibitem{burgess2019monet}
Christopher~P Burgess, Loic Matthey, Nicholas Watters, Rishabh Kabra, Irina
  Higgins, Matt Botvinick, and Alexander Lerchner.
\newblock Monet: Unsupervised scene decomposition and representation.
\newblock {\em arXiv preprint arXiv:1901.11390}, 2019.

\bibitem{dh2022}
Jean-Baptiste Camps, Chahan Vidal-Gorène, Dominique Stutzmann, Marguerite
  Vernet, and Ariane Pinche.
\newblock Data {Diversity} in handwritten text recognition: challenge or
  opportunity?
\newblock In {DH2022 Local Organizing Committee}, editor, {\em Digital
  {Humanities} 2022. {Conference} {Abstracts} ({The} {University} of {Tokyo},
  {Japan}, 25-29 {July} 2022)}, pages 160--165. Tokyo, 2022.

\bibitem{SPI-2}
Arianna Ciula.
\newblock Digital palaeography: using the digital representation of medieval
  script to support palaeographic analysis.
\newblock {\em Digital Medievalist}, 1, 2005.

\bibitem{crawford2019spatially}
Eric Crawford and Joelle Pineau.
\newblock Spatially invariant unsupervised object detection with convolutional
  neural networks.
\newblock In {\em Proceedings of the AAAI Conference on Artificial
  Intelligence}, volume~33, pages 3412--3420, 2019.

\bibitem{de2020htr}
Arthur~Flor de Sousa~Neto, Byron Leite~Dantas Bezerra, Alejandro~H{\'e}ctor
  Toselli, and Estanislau~Baptista Lima.
\newblock Htr-flor: a deep learning system for offline handwritten text
  recognition.
\newblock In {\em 2020 33rd SIBGRAPI Conference on Graphics, Patterns and
  Images (SIBGRAPI)}, pages 54--61. IEEE, 2020.

\bibitem{deng2020generative}
Fei Deng, Zhuo Zhi, Donghun Lee, and Sungjin Ahn.
\newblock Generative scene graph networks.
\newblock In {\em International Conference on Learning Representations}, 2020.

\bibitem{emami2021efficient}
Patrick Emami, Pan He, Sanjay Ranka, and Anand Rangarajan.
\newblock Efficient iterative amortized inference for learning symmetric and
  disentangled multi-object representations.
\newblock In {\em International Conference on Machine Learning}, pages
  2970--2981. PMLR, 2021.

\bibitem{eslami_attend_2016}
S.~M.~Ali Eslami, Nicolas Heess, Theophane Weber, Yuval Tassa, David
  Szepesvari, Koray Kavukcuoglu, and Geoffrey~E. Hinton.
\newblock Attend, {{Infer}}, {{Repeat}}: {{Fast Scene Understanding}} with
  {{Generative Models}}.
\newblock In {\em Advances in {{Neural Information Processing Systems}}},
  volume~29, Aug. 2016.

\bibitem{fogel_scrabblegan_2020}
Sharon Fogel, Hadar {Averbuch-Elor}, Sarel Cohen, Shai Mazor, and Roee Litman.
\newblock {{ScrabbleGAN}}: {{Semi-Supervised Varying Length Handwritten Text
  Generation}}.
\newblock {\em arXiv:2003.10557 [cs]}, Mar. 2020.

\bibitem{garrette_unsupervised_2015}
Dan Garrette, Hannah {Alpert-Abrams}, Taylor {Berg-Kirkpatrick}, and Dan Klein.
\newblock Unsupervised {{Code-Switching}} for {{Multilingual Historical
  Document Transcription}}.
\newblock In {\em Proceedings of the 2015 {{Conference}} of the {{North
  American Chapter}} of the {{Association}} for {{Computational Linguistics}}:
  {{Human Language Technologies}}}, pages 1036--1041, {Denver, Colorado}, 2015.
  {Association for Computational Linguistics}.

\bibitem{goodfellow2014generative}
Ian Goodfellow, Jean Pouget-Abadie, Mehdi Mirza, Bing Xu, David Warde-Farley,
  Sherjil Ozair, Aaron Courville, and Yoshua Bengio.
\newblock Generative adversarial nets.
\newblock {\em Advances in neural information processing systems}, 27, 2014.

\bibitem{ctc}
Alex Graves, Santiago Fern\'{a}ndez, Faustino Gomez, and J\"{u}rgen
  Schmidhuber.
\newblock Connectionist temporal classification: Labelling unsegmented sequence
  data with recurrent neural networks.
\newblock In {\em Proceedings of the 23rd International Conference on Machine
  Learning}, ICML '06, page 369–376, New York, NY, USA, 2006. Association for
  Computing Machinery.

\bibitem{graves2008offline}
Alex Graves and J{\"u}rgen Schmidhuber.
\newblock Offline handwriting recognition with multidimensional recurrent
  neural networks.
\newblock {\em Advances in neural information processing systems}, 21:545--552,
  2008.

\bibitem{greff2019multi}
Klaus Greff, Rapha{\"e}l~Lopez Kaufman, Rishabh Kabra, Nick Watters,
  Christopher Burgess, Daniel Zoran, Loic Matthey, Matthew Botvinick, and
  Alexander Lerchner.
\newblock Multi-object representation learning with iterative variational
  inference.
\newblock In {\em International Conference on Machine Learning}, pages
  2424--2433. PMLR, 2019.

\bibitem{greff2017neural}
Klaus Greff, Sjoerd Van~Steenkiste, and J{\"u}rgen Schmidhuber.
\newblock Neural expectation maximization.
\newblock {\em Advances in Neural Information Processing Systems}, 30, 2017.

\bibitem{gupta_learning_2018}
Ankush Gupta, Andrea Vedaldi, and Andrew Zisserman.
\newblock Learning to {{Read}} by {{Spelling}}: {{Towards Unsupervised Text
  Recognition}}.
\newblock {\em arXiv:1809.08675 [cs]}, Dec. 2018.

\bibitem{he2016deep}
Kaiming He, Xiangyu Zhang, Shaoqing Ren, and Jian Sun.
\newblock Deep residual learning for image recognition.
\newblock In {\em Proceedings of the IEEE conference on computer vision and
  pattern recognition}, pages 770--778, 2016.

\bibitem{hochberg1997automatic}
Judith Hochberg, Patrick Kelly, Timothy Thomas, and Lila Kerns.
\newblock Automatic script identification from document images using
  cluster-based templates.
\newblock {\em IEEE Transactions on Pattern Analysis and Machine Intelligence},
  19(2):176--181, 1997.

\bibitem{jaderberg_spatial_2015}
Max Jaderberg, Karen Simonyan, and Andrew Zisserman.
\newblock Spatial {{Transformer Networks}}.
\newblock In {\em Advances in {{Neural Information Processing Systems}}},
  volume~28, 2015.

\bibitem{jiang2020generative}
Jindong Jiang and Sungjin Ahn.
\newblock Generative neurosymbolic machines.
\newblock {\em Advances in Neural Information Processing Systems},
  33:12572--12582, 2020.

\bibitem{kahle2017transkribus}
Philip Kahle, Sebastian Colutto, G{\"u}nter Hackl, and G{\"u}nter
  M{\"u}hlberger.
\newblock Transkribus-a service platform for transcription, recognition and
  retrieval of historical documents.
\newblock In {\em 2017 14th IAPR International Conference on Document Analysis
  and Recognition (ICDAR)}, volume~4, pages 19--24. IEEE, 2017.

\bibitem{patwyr}
Lei Kang, Pau Riba, Mar{\c{c}}al Rusi{\~n}ol, Alicia Forn{\'e}s, and Mauricio
  Villegas.
\newblock Pay attention to what you read: Non-recurrent handwritten text-line
  recognition.
\newblock {\em arXiv preprint arXiv:2005.13044}, 2020.

\bibitem{karazija_clevrtex_2021}
Laurynas Karazija, Iro Laina, and Christian Rupprecht.
\newblock {{ClevrTex}}: {{A Texture-Rich Benchmark}} for {{Unsupervised
  Multi-Object Segmentation}}.
\newblock {\em arXiv:2111.10265 [cs]}, Nov. 2021.

\bibitem{knight_copiale_2011}
Kevin Knight, Beata Megyesi, and Christiane Schaefer.
\newblock The {{Copiale Cipher}}.
\newblock In {\em Proceedings of the {{ACL Workshop}} on {{Building}} and
  {{Using Comparable Corpora}}}, pages 2--9, 2011.

\bibitem{kopec1996document}
Gary~E Kopec and Mauricio Lomelin.
\newblock Document-specific character template estimation.
\newblock In {\em Document Recognition III}, volume 2660, pages 14--26. SPIE,
  1996.

\bibitem{kopec1997supervised}
Gary~E Kopec and Mauricio Lomelin.
\newblock Supervised template estimation for document image decoding.
\newblock {\em IEEE Transactions on Pattern Analysis and Machine Intelligence},
  19(12):1313--1324, 1997.

\bibitem{kopec2001n}
Gary~E Kopec, Maya~R Said, and Kris Popat.
\newblock N-gram language models for document image decoding.
\newblock In {\em Document Recognition and Retrieval IX}, volume 4670, pages
  191--202. SPIE, 2001.

\bibitem{lecun_backpropagation_1989}
Yann LeCun, Bernhard Boser, John~S Denker, Donnie Henderson, Richard~E Howard,
  Wayne Hubbard, and Lawrence~D Jackel.
\newblock Backpropagation applied to handwritten zip code recognition.
\newblock {\em Neural computation}, 1(4):541--551, 1989.

\bibitem{lecun_gradientbased_1998}
Yann LeCun, L{\'e}on Bottou, Yoshua Bengio, and Patrick Haffner.
\newblock Gradient-based learning applied to document recognition.
\newblock {\em Proceedings of the IEEE}, 86(11):2278--2324, 1998.

\bibitem{li2021trocr}
Minghao Li, Tengchao Lv, Lei Cui, Yijuan Lu, Dinei Florencio, Cha Zhang,
  Zhoujun Li, and Furu Wei.
\newblock Trocr: Transformer-based optical character recognition with
  pre-trained models.
\newblock {\em arXiv preprint arXiv:2109.10282}, 2021.

\bibitem{adamw}
Ilya Loshchilov and Frank Hutter.
\newblock Decoupled weight decay regularization.
\newblock {\em arXiv preprint arXiv:1711.05101}, 2017.

\bibitem{monnier_docextractor_2020}
Tom Monnier and Mathieu Aubry.
\newblock {{docExtractor}}: {{An}} off-the-shelf historical document element
  extraction.
\newblock In {\em 2020 17th {{International Conference}} on {{Frontiers}} in
  {{Handwriting Recognition}} ({{ICFHR}})}, pages 91--96, {Dortmund, Germany},
  Sept. 2020. {IEEE}.

\bibitem{monnier_deep_2020}
Tom Monnier, Thibault Groueix, and Mathieu Aubry.
\newblock Deep {{Transformation-Invariant Clustering}}.
\newblock In {\em {{NeurIPS}}}, Oct. 2020.

\bibitem{monnier_unsupervised_2021}
Tom Monnier, Elliot Vincent, Jean Ponce, and Mathieu Aubry.
\newblock Unsupervised {{Layered Image Decomposition}} into {{Object
  Prototypes}}.
\newblock In {\em Proceedings of the {{IEEE}}/{{CVF International Conference}}
  on {{Computer Vision}}}, pages 8640--8650, Apr. 2021.

\bibitem{nolan2010method}
Joseph~C Nolan and Robert Filippini.
\newblock Method and apparatus for creating a high-fidelity glyph prototype
  from low-resolution glyph images, Apr.~20 2010.
\newblock US Patent 7,702,182.

\bibitem{puig}
Joan Puigcerver.
\newblock Are multidimensional recurrent layers really necessary for
  handwritten text recognition?
\newblock In {\em 2017 14th IAPR International Conference on Document Analysis
  and Recognition (ICDAR)}, volume~1, pages 67--72. IEEE, 2017.

\bibitem{reddy2022search}
Pradyumna Reddy, Paul Guerrero, and Niloy~J Mitra.
\newblock Search for concepts: Discovering visual concepts using direct
  optimization.
\newblock {\em arXiv preprint arXiv:2210.14808}, 2022.

\bibitem{smirnov_marionette_2021}
Dmitriy Smirnov, Michael Gharbi, Matthew Fisher, Vitor Guizilini, Alexei~A.
  Efros, and Justin Solomon.
\newblock {{MarioNette}}: {{Self-Supervised Sprite Learning}}.
\newblock {\em arXiv:2104.14553 [cs]}, Apr. 2021.

\bibitem{HTRbyMatching}
Mohamed~Ali Souibgui, Alicia Forn{\'{e}}s, Yousri Kessentini, and Crina Tudor.
\newblock A few-shot learning approach for historical ciphered manuscript
  recognition.
\newblock {\em CoRR}, abs/2009.12577, 2020.

\bibitem{pal}
Dominique. Stutzmann.
\newblock Fontenay dataset. original charters from fontenay before 1213
  \url{https://doi.org/10.5281/zenodo.6507963}.

\bibitem{vaswani2017attention}
Ashish Vaswani, Noam Shazeer, Niki Parmar, Jakob Uszkoreit, Llion Jones,
  Aidan~N Gomez, {\L}ukasz Kaiser, and Illia Polosukhin.
\newblock Attention is all you need.
\newblock {\em Advances in neural information processing systems}, 30, 2017.

\bibitem{vincent_google_2007}
L. Vincent.
\newblock Google {{Book Search}}: {{Document Understanding}} on a {{Massive
  Scale}}.
\newblock In {\em Ninth {{International Conference}} on {{Document Analysis}}
  and {{Recognition}} ({{ICDAR}} 2007) {{Vol}} 2}, pages 819--823, {Curitiba,
  Parana, Brazil}, Sept. 2007. {IEEE}.

\bibitem{xu1999prototype}
Yihong Xu and George Nagy.
\newblock Prototype extraction and adaptive ocr.
\newblock {\em IEEE Transactions on Pattern Analysis and Machine Intelligence},
  21(12):1280--1296, 1999.

\bibitem{yang2020learning}
Yanchao Yang, Yutong Chen, and Stefano Soatto.
\newblock Learning to manipulate individual objects in an image.
\newblock In {\em Proceedings of the IEEE/CVF Conference on Computer Vision and
  Pattern Recognition}, pages 6558--6567, 2020.

\bibitem{vicky-unet}
Vickie Ye, Zhengqi Li, Richard Tucker, Angjoo Kanazawa, and Noah Snavely.
\newblock Deformable sprites for unsupervised video decomposition.
\newblock In {\em IEEE Conference on Computer Vision and Pattern Recognition
  (CVPR)}, June 2022.

\bibitem{zhang_adaptive_2020}
Chuhan Zhang, Ankush Gupta, and Andrew Zisserman.
\newblock Adaptive {{Text Recognition}} through {{Visual Matching}}.
\newblock {\em ECCV 2020}, Sept. 2020.

\end{thebibliography}
}

\end{document}